\documentclass[12pt,verbose=true,letterpaper,margin=1.5in]{article}
\usepackage{arxiv}

\newcommand{\ignore}[1]{}
\usepackage[utf8]{inputenc}
\usepackage[vlined,linesnumbered,ruled,titlenotnumbered]{algorithm2e}
\usepackage{amsmath,amsthm}

\usepackage{dsfont}

\usepackage{multirow, rotating}

\usepackage{adjustbox}

\DeclareMathOperator{\rand}{rand}

\begin{document}

\author{Aneta Neumann
\\Optimisation and Logistics\\
School of Computer Science
\\The University of Adelaide
\\Adelaide, Australia
\And
Denis Antipov
\\ITMO University
\\St. Petersburg, Russia
\And
Frank Neumann
\\Optimisation and Logistics\\
School of Computer Science
\\The University of Adelaide
\\Adelaide, Australia
}

\title{Coevolutionary Pareto Diversity Optimization}

\maketitle

\begin{abstract}
  Computing diverse sets of high quality solutions for a given optimization problem has become an important topic in recent years. In this paper, we introduce a coevolutionary Pareto Diversity Optimization approach which builds on the success of reformulating a constrained single-objective optimization problem as a bi-objective problem by turning the constraint into an additional objective. Our new Pareto Diversity optimization approach uses this bi-objective formulation to optimize the problem while also maintaining an additional population of high quality solutions for which diversity is optimized with respect to a given diversity measure. We show that our standard co-evolutionary Pareto Diversity Optimization approach outperforms the recently introduced DIVEA algorithm which obtains its initial population by generalized diversifying greedy sampling and improving the diversity of the set of solutions afterwards. Furthermore, we study possible improvements of the Pareto Diversity Optimization approach. In particular, we show that the use of inter-population crossover further improves the diversity of the set of solutions.
\end{abstract}


\section{Introduction}

The classical goal in the area of single-objective optimization is to compute a single optimal or near optimal solution for the given problem. In the case of multi-objective optimization, there is usually a conflicting set of objectives and the goal is to compute the set of trade-offs of the given objective functions.

It has been shown in various studies that the use of bi-objective problem formulations of a constrained single-objective optimization problem is often highly beneficial to the search process of evolutionary algorithms. The use of evolutionary multi-objective algorithms to tackle a constrained single-objective optimization problem by using the constrained cost function as an additional objective enables an evolutionary algorithm to mimic greedy behavior. This often leads to provably optimal solutions or good approximations for a wide range of optimization problems~\cite{DBLP:journals/nc/NeumannW06,DBLP:journals/ec/FriedrichHHNW10,DBLP:journals/algorithmica/KratschN13}. 

This approach is recently often referred to as Pareto optimization and has been shown to obtain high quality solutions for a wide range of machine learning problems~\cite{DBLP:books/sp/ZhouYQ19}. Due to its problem formulation, this approach provides the user with a set of solutions representing the trade-offs with respect to the objective function of the single-objective optimization problem and its constrained function. 
Pareto optimization approaches have been shown to achieve good approximations for a wide range of submodular optimization problems~\cite{DBLP:journals/ec/FriedrichN15,DBLP:conf/ijcai/QianSYT17,DBLP:conf/nips/QianYZ15,DBLP:conf/ppsn/NeumannN20}. Again the main underlying reason is that they are able to mimic greedy approaches which provide best worst-case performance guarantees for many monotone submodular optimization problems~\cite{DBLP:journals/mp/NemhauserWF78,DBLP:journals/ipl/KhullerMN99,DBLP:conf/aaai/ZhangV16}.
Approaches are usually analyzed and experimentally investigated for subset selection problems where solutions are constructed by selecting elements from a given set of elements. This leads to the natural encoding of possible solutions as binary strings. We will use that representation throughout this paper. However, it should be noted that the approach can easily be transferred to problems working with other problem encoding.

Another area that has recently gained significant interest in the evolutionary algorithm literature is the computation of high quality diverse sets of solutions for single-objective optimization problems~\cite{DBLP:conf/gecco/UlrichT11,doi:10.1162/evcoa00274}. In the case of robotics and games various types of quality diversity (QD) algorithms have been developed~\cite{DBLP:conf/gecco/PughSSS15,DBLP:conf/ppsn/PughSS16}. In the context of combinatorial optimization, evolutionary diversity optimization approaches have been designed to compute sets of problem instances~\cite{doi:10.1162/evcoa00274,DBLP:conf/gecco/NeumannGDN018,DBLP:conf/gecco/NeumannG0019} that can differentiate between the performance of two algorithms on such problems as the traveling salesperson problem. Furthermore, recently this approach has been used  to compute sets of high quality solutions of large diversity for classical combinatorial optimization problems such as the knapsack problem~\cite{BossekN021KP} and the traveling salesperson problem~\cite{viet2020evolving,DBLP:conf/gecco/NikfarjamBN021,DBLP:conf/foga/NikfarjamB0N21}. Furthermore, for constrained monotone submodular functions an evolutionary diversity optimization approach called DIVEA has been developed in \cite{DBLP:conf/gecco/NeumannB021} which improves an initial diverse population obtained by a diversifying greedy sampling technique.

Diversity plays a crucial role in the area of evolutionary multi-objective optimization. Evolutionary multi-objective algorithms aim to achieve a set of solutions that is close to the Pareto front and has significant diversity such that the Pareto front is covered well. Apart from classical diversity mechanisms such as computing the crowding distance in NSGA-II~\cite{DBLP:journals/tec/DebAPM02}, diversity has also been used as an additional objective in evolutionary multi-objective algorithms. It has been shown in \cite{DBLP:journals/ec/ToffoloB03} that using diversity as an additional objective can lead to better results in evolutionary multi-objective optimization. In \cite{DBLP:conf/cec/SeguraCSML13} diversity mechanisms have been introduced to improve the quality of solutions when applying Pareto optimization to obtain a single high quality solution for a single-objective optimization problem. Furthermore, niching~\cite{DBLP:conf/cec/WessingPR13} and diversity driven selection methods~\cite{DBLP:conf/emo/CarranoCT21} have been introduced to improve the results of evolutionary multi-objective algorithms. All these results either aim for a single solution when using Pareto optimization for single-objective problems or try to obtain better approximations of the Pareto front. In contrast to this, our approach aims to compute a diverse set of high quality solutions for a given single-objective problem.

With this paper, we bridge the areas of Pareto optimization, which in its standard approach computes a single high quality solution for a given single-objective optimization problem, and evolutionary diversity optimization. We construct coevolutionary Pareto Optimization approaches that can (1) maximize the quality of solutions and (2) maximize the diversity among high quality solutions for the given single-objective problem. The key idea is to have a population $P_O$ for the optimization part and a population $P_D$ for the diversity optimization part coevolving and interacting with each other. In our baseline Pareto Diversity Optimization (PDO) approach the interacting is only through the possibility of integrating an offspring obtained by mutation in both populations. Our advanced Pareto diversity optimization approaches use an inter-population based crossover operator which allows to combine individuals within a population but also between populations. 

We experimentally examine our approaches and compare them against DIVEA~\cite{DBLP:conf/gecco/NeumannB021} on the NP-hard submodular maximum coverage problem in graphs.
The maximum coverage problem is a well-known submodular optimization problem and has been widely used as a benchmark problem to test the performance of algorithms for the wide class of submodular problems. We investigate different instances of this problem together with different settings for the diversity optimization component. In particular, we are interested in how the different in population size and quality requirement impacts the performance of the algorithms. A wide range of parameter combinations are investigated in our experimental studies to make sure that our findings a widely applicable.
Our results show that our baseline Pareto optimization approach significantly outperforms DIVEA both in terms of the quality of the best solution that is achieved and the diversity of the set of solutions for almost all instance and parameter combinations. Our advanced Pareto optimization approaches show a further increase in the diversity of the set of solution while obtaining solutions of the same quality as the baseline PDO approach.

The paper is structured as follows. In Section~\ref{sec2}, we introduce the concepts of Pareto optimization and evolutionary diversity optimization. Section~\ref{sec3} provides our baseline Pareto diversity optimization algorithms and improvements are presented in Section~\ref{sec4}. We carry out experimental investigations that compare Pareto diversity optimization to DIVEA in Section~\ref{sec5}. Finally, we finish with some concluding remarks.

\section{Preliminaries}
\label{sec2}

We first consider subset selection problems with a monotone objective function with a single monotone cost constraint. 
Let $V=\{v_1, \ldots, v_n\}$ a set of $n$ input elements. We consider the search space $\{0,1\}^n$ and for a given solution $x \in \{0,1\}^n$, element $v_i$ is selected iff $x_i=1$ holds.

Let $f \colon \{0,1\}^n \rightarrow \mathds{R}_{\geq 0}$ be the given monotone objective function taking on only non-negative values, $c \colon \{0,1\}^n \rightarrow \mathds{R}_{\geq 0}$ be the monotone non negative cost function, and $B$ be the cost constraint. A solution $x$ is feasible iff $c(x) \leq B$ holds.
The classical goal is to compute a solution $x^{opt} = \arg \max_{x \in \{0,1\}^n \mid c(x) \leq B} f(x)$.

Our aim is computing a set of solutions $P_D$ where all solutions are feasible and meet a given quality threshold. Furthermore, we would like to maximize the quality of the best solution $x^*$ in $P_D$ and obtain a good approximation of the optimal solution $x^{opt}$.

\subsection{Evolutionary Diversity Optimization}

Evolutionary diversity optimization (EDO) aims to create a diverse set of solutions where all solutions meet a given quality criterion.
EDO evolves a set of solutions $P_D = \{x^1, \ldots, x^{\mu}\}$ where all solutions are feasible, i.e. $c(x) \leq B$ holds and meet a given quality thresholds value. As done in \cite{DBLP:conf/gecco/NeumannB021}, we require $f(x) \geq f_{\min}$ for all $x \in P_D$ where $f_{\min}$ is a thresholds value on the quality of solutions in $P_D$.

Let $V=\{v_1, \ldots, v_n\}$ be the given set of input elements.
For a given population $P$ of size $\mu$, we denote by
\begin{align*}
p(v_i) = \frac{|\{x^j \in P\mid v_i \in x^j\}|}{\mu}
\end{align*}
the fraction of solutions in $P$ that contain element $v_i$, $1 \leq i \leq n$.
We define the entropy of a given population $P$ as
$$
H(P) = - \sum_{i=1}^n p(v_i) \log_2 p(v_i),
$$
where for all $i$  such that $p(v_i) = 0$ we assume that $p(v_i) \log_2 p(v_i) = 0$.
Our goal is to maximise the entropy under the condition that all solutions in $P$ are feasible and meet a given quality criterion.
We use $\log(x)$ instead of $\log_2(x)$ in the following to simplify notations.

We consider an elitist version of the DIVEA (see Algorithm~\ref{alg:divea}) which has been introduced in~\cite{DBLP:conf/gecco/NeumannB021} for the optimization of constrained monotone submodular functions. The algorithm starts with a population that is obtained by generalized diversifying greedy sampling (GDGS). GDGS runs the generalized greedy algorithm for the given problem with budget $B-m$ (instead of $B$), where $m \geq 0$ is a given margin, and introduces additional randomly chosen elements as long as the constraint bound $B$ is not violated. In total $\mu$ individuals are created in this way.  For details on the GDGS, we refer the reader to~\cite{DBLP:conf/gecco/NeumannB021}. 

The population obtained by GDGS is the initial population for DIVEA which is then improved in terms of its diversity.
This is done by applying a $(\mu+1)$ evolutionary algorithm approach that only accepts feasible solutions meeting the quality threshold $f_{\min}$. In the selection step, the individual with the smallest contribution to diversity is removed.
 The only difference to the algorithm introduced in \cite{DBLP:conf/gecco/NeumannB021} is the additional use of elitism which makes sure that an individual $x^*$ (see line 9) of highest quality can not be removed from $P_D$ if the population size needs to be reduced. We carried out that change as a non elitist version of DIVEA would be less competitive in terms of the quality of the best solution that is produced. We compare this version of DIVEA to our Pareto Diversity optimization approaches later on.

\begin{algorithm}[t]
Input: Initial population $P_D$ with $|P_D|=\mu$ obtained by generalized diversifying greedy sampling.\;
$f_{min}= \min_{x \in P_D} f(x)$\;
$t \leftarrow 0$\;
\While{$t < t_{max}$}{
$t \leftarrow t+1$\;
Choose $x \in P$ uniformly at random and produce an offspring $y$ of $x$ by mutation\;
\If{$(f(y) \geq f_{min}) \wedge (c(y) \leq B)$}{ $P \leftarrow P \cup \{y\}$\;
$x^*= \arg \max_{x \in P_D} g_1(x)$\;
  $\hat{x} = \arg \max_{x \in (P_D \setminus\{x^*\})} H(P_D \setminus \{x\})$\;
  $P_D \leftarrow P_D \setminus \{\hat{x}\}$\;}
}
 \Return{$P_D=\{P_1, \ldots, P_{\mu}\}$}\;
\caption{Elitist Diversifying EA (DIVEA)}
\label{alg:divea}
\end{algorithm}

\subsection{Pareto Optimization}
Multi-objective formulations of constrained single-objective optimization have been shown to be highly beneficial both from a theoretical and practical perspective. Recently named as Pareto optimization, these approaches optimize the given goal function and the slack in the constraint at the same time. 
For our investigations, we follow the Pareto optimization approach and setup given in \cite{DBLP:conf/ijcai/QianSYT17}.

Each search point $x \in \{0,1\}^n$ is assigned an objective vector $g(x) = (g_1(x), g_2(x))$ where we have

\begin{equation}
    g_1(x) = \begin{cases}
    f(x) \text{ iff } c(x) \leq B+1\\
    -1 \text{ iff } c(x) > B+1
    \end{cases}
\end{equation}
and $g_2(x) = c(x)$. The aim is to maximize $g_1$ and minimize $g_2$ at the same time.
A search $y$ weakly dominates a search point z (denoted as $y \succeq z$ if $g_1(y) \geq g_1(z)$ and $g_2(y) \leq g_2(z)$ holds. Furthermore $y$ strongly dominates $z$ (denoted as $y \succ z$) iff $y \succeq z$ and $g(y)\not=g(z)$.

Such a setup is known to enable a greedy behavior if the algorithm stores at any point in time the set of non-dominated solutions obtained so far. This is highly beneficial for the design of evolutionary algorithms with approximation guarantees when dealing with monotone submodular optimization problems and a wide range of such results have been obtained~\cite{DBLP:journals/ec/FriedrichN15,DBLP:conf/ijcai/QianSYT17,DBLP:conf/nips/QianYZ15,DBLP:books/sp/ZhouYQ19}. In the following, we will show how to transfer such Pareto Optimization approaches into Pareto Diversity Optimization approaches which are not only able to compute a single high quality solution but a set of high quality solutions.

\section{Coevolutionary Pareto Diversity Optimization}
\label{sec3}
Our coevolutionary Pareto Diversity Optimization approach (PDO for brevity) is shown in Algorithm~\ref{alg:CO-EA}.
We work with two populations $P_O$ and $P_D$ that are interacting with each other.
The population $P_O$ is used to carry out standard Pareto optimization with bound $B$ whereas $P_D$ is used to evolve a set of solution of size at most $\mu$ where for all $x \in P_D$ $f(x) \geq f_{\min}$ and $c(x) \leq B$ holds.
The parameter $f_{\min}$ is the threshold on the quality of solutions that are used for diversity optimization using the population $P_D$. We determine this parameter as in DIVEA based on GDGS but make no use of the population obtained by greedy sampling.

\begin{algorithm}[t]
Input: $f_{\min}$\;
Choose $x \in \{0,1\}^n$ uniformly at random\;
$P_O \leftarrow \{x\}$\;
$t=0$\;
\While{$t < t_{\max}$}{
$t \leftarrow t+1$\;
\If{$|P_D|>1$ and $\rand() <0.5$}{
Choose $x \in P_D$ uniformly at random\;}
\Else{Choose $x \in P_O$ uniformly at random\;}
Create $y$ from $x$ by mutation\;
Determine $g(y)$\;

\If{$\not\exists w \in P_O: w \succ y$} {
  $P_O \leftarrow (P_O \cup \{y\})\backslash \{z\in P_O \mid y \succeq z\}$\;
}
  \If{$g_1(y) \geq f_{\min} \wedge g_2(y) \leq B$}{
  $P_D \leftarrow P_D \cup \{y\}$\;

  \If{$|P_D| > \mu$}{
  $x^*= \arg \max_{x \in P_D} g_1(x)$\;
  $\hat{x} = \arg \max_{x \in (P_D \setminus\{x^*\})} H(P_D \setminus \{x\})$\;
  $P_D \leftarrow P_D \setminus \{\hat{x}\}$\;
  }
  }

  }
\caption{Pareto Diversity Optimization (PDO)} \label{alg:CO-EA}
\end{algorithm}

PDO works as follows. We first follow the standard Pareto optimization algorithms and produce in each iteration one offspring by first choosing its population (each with probability $0.5$ or, if $P_D$ is empty, we choose $P_O$) and then choosing an individual from that population uniformly at random. Then we apply a mutation operator to the chosen individual. Then we try to add this individual into both populations as follows.

Consider $P_O$ and $P_D$ and let $y$ be an offspring created in the current iteration.
For the population $P_O$, we apply standard dominance selection, i.e. we include $y$ in $P_O$ if there is no individual $w \in P_O$ which strictly dominates $y$ (denoted as $w \succ y$). If $y$ is included in $P_O$, then each individual $z \in P_O$ that is weakly dominated by $y$ (denoted as $y \succeq z$) is removed from $P_O$. This approach lets us keep the set of Pareto-optimal solutions in $P_O$. 

For $P_D$ we use the following procedure.
If $f(y) > f_{\min}$ and $c(x) \leq B$ then we add $y$ to $P_D$. In order to make sure that $|P_D|\leq \mu$ in each iteration, we remove an individual from $P_D$ if $|P_D|>\mu$ holds after $y$ has been added. Let $x^*$ be a (feasible) individual in $P_D$ that has the largest function value. As our approach should maximize quality and diversity, we remove an individual from $P_D$ that has the smallest contribution to diversity among all individuals in $P_D \setminus \{x^*\}$. Note that $x^*$ will not be removed which ensures an elitist behavior with respect to the objective function $f$ in $P_D$. Furthermore, the currently best feasible solution of fitness at least $f_{\min}$ is always added to $P_D$ which implies that the best solution obtained during the run of the algorithm is contained in $P_D$ once a solution $x$ with $f(x) \geq f_{\min}$ has been obtained for the first time.

The coevolution of $P_D$ and $P_O$ is supposed to speed up the diversity growth in $P_D$. Since the selection mechanism in $P_O$ is different from the one in $P_D$, these two populations are likely to have quite different sets of individuals in terms of genotype, hence feasible offspring created from individuals of $P_O$ are likely to increase the diversity in $P_D$.

\section{Improvements based on inter-population crossover}
\label{sec4}

PDO works with two populations $P_O$ and $P_D$ and the standard approach outlined in the previous section uses an interaction between these populations as a part of the selection steps which investigates the suitability of a produced offspring regarding its suitability for both $P_O$ and $P_D$. In this section we show a modification of PDO which uses crossover to create a new individual and which we call PDO-C.

\begin{algorithm}[t]
Input: Populations $P_O$ and $P_D$ with $|P_O|>1$ \;

\If{$|P_D|\ge 1$ and $\rand() <0.5$}{
Choose $s \in P_D$ uniformly at random\;
\If{$|P_D| + |P_O| \ge 2$ and $\rand() < p_c$}{
\If{$|P_D|\ge 2$ and $\rand() <0.5$}{
Choose $t \in (P_D \setminus\{s\})$ uniformly at random\;
}\Else{
Choose $t \in P_O$ uniformly at random\;}
s = Crossover(s,t)\;
y = mutation(s)\;
}
\Else{
y = mutation(s)\;
}
}
\Else{
Choose $s \in P_O$ uniformly at random\;
\If{$|P_D| + |P_O| \ge 2$ and $\rand() < p_c$}{
\If{$|P_D|\ge 1$ and $\rand() < 0.5$}{
Choose $t \in P_D$ uniformly at random\;
}\ElseIf{$|P_O|\ge 2$}{
Choose $t \in (P_O \setminus\{s\})$ uniformly at random\;}
s = Crossover(s,t)\;
y = mutation(s)\;
}
\Else{
y = mutation(s)\;
}
}

\caption{Inter-population crossover and mutation for Pareto Diversity Optimization}
\label{alg:crossmut}
\end{algorithm}

\begin{algorithm}[t]
Input: Solution $x$ (with $c(x)> B$)\;
\While{$c(x) >B$}{
Remove an element of $x$ by flipping a randomly chosen $1$-bit of $x$.
}
\caption{Repair operator}
\label{alg:repair}
\end{algorithm}

We now introduce an inter-population crossover and mutation approach shown in Algorithm~\ref{alg:crossmut}, which we use in PDO-C to create a new individual. The crossover operator is applied with probability $p_c$ (but only when we already have at least two individuals, that is, $|P_D| + |P_O| \ge 2$) and picks the first individual $s$ from either $P_O$ or $P_D$ and may select the second individual $t$ also either from $P_O$ or $P_D$.

More precisely, we first choose a population for selecting $s$, $P_O$ or $P_D$, each with probability $0.5$ (if $P_D$ is empty, then we choose $P_O$) and then we choose $s$ from that population uniformly at random. Then for $t$ we also select one of the populations equiprobably, but we cannot choose an empty population, or the population which consisted only of the first chosen parent $s$. Then we choose $t$ uniformly at random from the chosen population (but we ensure not to choose $s$ as the second parent). 

In this way crossover operations within a population and between the populations are possible. After an offspring $y$ has been created by crossover, this individual is also undergoing a mutation operation. With probability $1 - p_c$ we do not perform crossover and produce $y$ by selecting an individual $s$ in the same way as in PDO (Algorithm~\ref{alg:CO-EA}) and mutating it.

As crossover operations are often leading to infeasible solutions, a repair operator (shown in Algorithm~\ref{alg:repair}) is applied after the offspring generation by crossover and mutation. This is done by removing randomly chosen elements until the constraint bound is met.
It should be noted that different crossover operators can be used to produce the offspring $y$ from $s$ and $t$. For our investigations, we use uniform crossover which is one of the standard crossover operators when working with binary strings. We call this modification using crossover PDO-C.

We also consider a further modification of PDO-C which uses a heavy tail mutation~\cite{DBLP:conf/gecco/DoerrLMN17} instead of the standard bit mutation used by PDO and PDO-C. We call it PDO-CH. 
In each application of the heavy tail mutation operator, first a parameter $\alpha \in [1..n/2]$ is chosen according to the discrete power law distribution $D_{n/2}^{\beta}$. Afterwards, each bit is flipped with probability $\alpha/n$. This allows to flip significantly more bits in some mutation steps than the use of standard bit mutations, while preserving a constant probability to flip exactly one bit. This way of mutation is supposed to improve the exploration-via-mutation ability of the algorithm, while not harming the exploitation-via-mutation.

From a theoretical perspective, the use of heavy tail mutations has been shown to be effective on the OneMax benchmark problem in~\cite{DBLP:conf/gecco/DoerrLMN17,DBLP:conf/gecco/AntipovBD20}. 
Furthermore, it has been shown in \cite{DBLP:conf/gecco/XieN020} that the use of heavy tail mutation further improves the performance in a Pareto optimization set up which already includes crossover for the chance constrained knapsack problem.

\begin{table*}[]

    \centering
    \caption{Optimization and Diversity Results for DIVEA and PDO}
    \label{tab:PDO}
    \begin{scriptsize}
    \begin{tabular}{|c|c|c|c||c|c|c|c|c||c|c|c|c|c|c|}
    \hline
    \multirow{2}{*}{Graph} & \multirow{2}{*}{$B$} & \multirow{2}{*}{$m$} & \multirow{2}{*}{$\mu$} & \multicolumn{5}{|c|}{Optimization} &  \multicolumn{3}{|c|}{Diversity} \\
    
  & &   & & GDGS$_w$ & GDGS$_B$ & DIVEA  & PDO  & $p_1$ & DIVEA  & PDO   & $p_2$\\ \hline
\multirow{16}{*}{frb30-15-1-mis}&  20000 & 2000 & 10 & 276.20 & 281.83 & 299.37 &  \textbf{303.60} & 0.000 & 15.34 & \textbf{16.92} & 0.000 \\
 & 20000 & 2000 & 20 & 276.00 & 282.00 & 298.40 & \textbf{303.70} & 0.000 & 17.25 & \textbf{19.11} & 0.000 \\
 & 20000 & 2000 & 50 & 276.00 & 282.00 & 298.13 &  \textbf{303.50} & 0.000 & 18.23 & \textbf{20.59} & 0.000 \\
 & 20000 & 2000 & 100 & 276.00 & 282.00 & 297.60 &  \textbf{303.40} & 0.000 & 18.50 & \textbf{21.10} & 0.000 \\
 & 20000 & 4000 & 10 & 266.67 & 288.73 & 299.50 &  \textbf{303.80} & 0.000 & 17.53 & \textbf{17.97} & 0.003 \\
 & 20000 & 4000 & 20 & 265.53 & 291.00 & 298.33 &  \textbf{303.77} & 0.000 & 20.11 & \textbf{20.73} & 0.000 \\
 & 20000 & 4000 & 50 & 263.53 & 294.97 & 297.97 &  \textbf{303.63} & 0.000 & 21.84 & \textbf{23.03} & 0.000 \\
 & 20000 & 4000 & 100 & 262.80 & 296.07 & 297.90 &  \textbf{303.60} & 0.000 & 22.32 & \textbf{23.81} & 0.000 \\ 
 & 40000 & 2000 & 10 & 403.00 & 405.00 & 409.03 &  \textbf{413.83} & 0.000 & 15.75 & \textbf{24.12} & 0.000 \\
 & 40000 & 2000 & 20 & 403.00 & 405.00 & 408.60 &  \textbf{414.10} & 0.000 & 16.66 & \textbf{26.51} & 0.000 \\
 & 40000 & 2000 & 50 & 403.00 & 405.00 & 408.63 &  \textbf{413.93} & 0.000 & 16.54 & \textbf{27.86} & 0.000 \\
 & 40000 & 2000 & 100 & 403.00 & 405.00 & 408.13 &  \textbf{414.23} & 0.000 & 16.68 & \textbf{28.51} & 0.000 \\
 & 40000 & 4000 & 10 & 394.73 & 403.23 & 407.77 &  \textbf{414.17} & 0.000 & 25.09 & \textbf{27.77} & 0.000 \\
 & 40000 & 4000 & 20 & 394.33 & 404.47 & 407.43 &  \textbf{413.90} & 0.000 & 28.62 & \textbf{30.94} & 0.000 \\
 & 40000 & 4000 & 50 & 392.73 & 405.20 & 406.77 &  \textbf{413.97} & 0.000 & 31.09 & \textbf{33.63} & 0.000 \\
 & 40000 & 4000 & 100 & 392.70 & 405.63 & 406.97 &  \textbf{413.50} & 0.000 & 31.55 & \textbf{34.14} & 0.000 \\ \hline
 \multirow{16}{*}{frb30-15-2-mis}  & 20000 & 2000 & 10 & 257.00 & 257.00 & 286.40 & \textbf{289.27} & 0.000 & 15.04 & \textbf{16.38} & 0.000 \\
 & 20000 & 2000 & 20 & 257.00 & 257.00 & 285.63 & \textbf{289.17} & 0.000 & 17.31 & \textbf{19.11} & 0.000 \\
 & 20000 & 2000 & 50 & 257.00 & 257.00 & 286.03 & \textbf{289.07} & 0.000 & 19.35 & \textbf{21.34} & 0.000 \\
 & 20000 & 2000 & 100 & 257.00 & 257.00 & 284.17 & \textbf{289.23} & 0.000 & 19.69 & \textbf{22.27} & 0.000 \\
 & 20000 & 4000 & 10 & 256.00 & 279.53 & 286.10 & \textbf{289.33} & 0.000 & 15.41 & \textbf{16.47} & 0.000 \\
 & 20000 & 4000 & 20 & 255.00 & 281.87 & 286.37 & \textbf{289.10} & 0.000 & 18.06 & \textbf{19.29} & 0.000 \\
 & 20000 & 4000 & 50 & 253.07 & 283.77 & 285.60 & \textbf{289.17} & 0.000 & 20.73 & \textbf{21.73} & 0.000 \\
 & 20000 & 4000 & 100 & 251.63 & 284.77 & 285.73 & \textbf{289.07} & 0.000 & 21.94 & \textbf{22.73} & 0.000 \\
 & 40000 & 2000 & 10 & 396.27 & 402.67 & 409.63 & \textbf{415.10} & 0.000 & 23.15 & \textbf{25.11} & 0.000 \\
 & 40000 & 2000 & 20 & 396.07 & 403.00 & 407.60 & \textbf{415.47} & 0.000 & 25.60 & \textbf{27.56} & 0.000 \\
 & 40000 & 2000 & 50 & 396.00 & 403.00 & 407.53 & \textbf{415.57} & 0.000 & 27.26 & \textbf{29.34} & 0.000 \\
 & 40000 & 2000 & 100 & 396.00 & 403.00 & 407.43 & \textbf{415.63} & 0.000 & 27.66 & \textbf{30.15} & 0.000 \\
 & 40000 & 4000 & 10 & 392.13 & 400.43 & 409.40 & \textbf{415.37} & 0.000 & 25.12 & \textbf{26.23} & 0.000 \\
 & 40000 & 4000 & 20 & 391.60 & 401.00 & 407.97 & \textbf{415.03} & 0.000 & 28.46 & \textbf{29.56} & 0.000 \\
 & 40000 & 4000 & 50 & 391.20 & 402.03 & 406.97 & \textbf{415.67} & 0.000 & 30.20 & \textbf{31.56} & 0.000 \\
 & 40000 & 4000 & 100 & 391.00 & 402.60 & 406.40 & \textbf{415.17} & 0.000 & 30.64 & \textbf{32.16} & 0.000 \\ \hline

 \multirow{16}{*}{frb35-17-1-mis} & 20000 & 2000 & 10 & 274.40 & 296.67 & 299.03 & \textbf{300.00} & 0.000 & 10.95 & \textbf{14.99} & 0.000 \\
 & 20000 & 2000 & 20 & 274.13 & 297.60 & 298.83 & \textbf{300.00} & 0.000 & 12.26 & \textbf{17.15} & 0.000 \\
 & 20000 & 2000 & 50 & 274.00 & 298.00 & 298.50 & \textbf{300.00} & 0.000 & 12.52 & \textbf{18.58} & 0.000 \\
 & 20000 & 2000 & 100 & 274.00 & 298.00 & 298.67 & \textbf{300.00} & 0.000 & 12.76 & \textbf{19.05} & 0.000 \\
 & 20000 & 4000 & 10 & 268.70 & 294.33 & 297.87 & \textbf{300.00} & 0.000 & 12.25 & \textbf{15.44} & 0.000 \\
 & 20000 & 4000 & 20 & 267.87 & 295.73 & 297.00 & \textbf{300.00} & 0.000 & 15.36 & \textbf{17.93} & 0.000 \\
 & 20000 & 4000 & 50 & 267.00 & 296.00 & 297.20 & \textbf{300.00} & 0.000 & 15.77 & \textbf{19.50} & 0.000 \\
 & 20000 & 4000 & 100 & 267.00 & 296.00 & 296.67 & \textbf{300.00} & 0.000 & 16.86 & \textbf{19.96} & 0.000 \\
 & 40000 & 2000 & 10 & 429.00 & 429.00 & 452.57 & \textbf{465.20} & 0.000 & 24.22 & \textbf{25.35} & 0.000 \\
 & 40000 & 2000 & 20 & 429.00 & 429.00 & 451.03 & \textbf{465.40} & 0.000 & 26.91 & \textbf{28.40} & 0.000 \\
 & 40000 & 2000 & 50 & 429.00 & 429.00 & 450.60 & \textbf{465.57} & 0.000 & 28.12 & \textbf{30.11} & 0.000 \\
 & 40000 & 2000 & 100 & 429.00 & 429.00 & 450.23 & \textbf{464.77} & 0.000 & 29.01 & \textbf{30.72} & 0.000 \\
 & 40000 & 4000 & 10 & 427.33 & 443.50 & 451.87 & \textbf{465.10} & 0.000 & 24.88 & \textbf{25.40} & 0.021 \\
 & 40000 & 4000 & 20 & 426.10 & 443.10 & 450.87 & \textbf{465.40} & 0.000 & 28.48 & \textbf{28.83} & 0.107 \\
 & 40000 & 4000 & 50 & 425.40 & 444.97 & 449.30 & \textbf{465.33} & 0.000 & 30.45 & \textbf{31.03} & 0.002 \\
 & 40000 & 4000 & 100 & 425.13 & 445.60 & 449.67 & \textbf{465.37} & 0.000 & 30.75 & \textbf{31.75} & 0.000 \\ \hline
 \multirow{16}{*}{frb40-19-1-mis} & 20000 & 2000 & 10 & 239.00 & 239.00 & 269.10 & \textbf{272.50} & 0.000 & 12.31 & \textbf{12.95} & 0.000 \\
 & 20000 & 2000 & 20 & 239.00 & 239.00 & 269.43 & \textbf{272.50} & 0.000 & 14.75 & \textbf{15.24} & 0.000 \\
 & 20000 & 2000 & 50 & 239.00 & 239.00 & 268.40 & 272.50 & 0.000 & 15.96 & 16.55 & 0.000 \\
 & 20000 & 2000 & 100 & 239.00 & 239.00 & 268.70 & \textbf{272.47} & 0.000 & 16.31 & \textbf{16.97} & 0.000 \\
 & 20000 & 4000 & 10 & 239.00 & 239.00 & 270.00 & \textbf{272.37} & 0.000 & 12.49 & \textbf{12.96} & 0.000 \\
 & 20000 & 4000 & 20 & 239.00 & 239.00 & 269.17 & \textbf{272.43} & 0.000 & 14.78 & \textbf{15.21} & 0.000 \\
 & 20000 & 4000 & 50 & 239.00 & 239.00 & 268.90 & \textbf{272.47} & 0.000 & 15.91 & \textbf{16.55} & 0.000 \\
 & 20000 & 4000 & 100 & 239.00 & 239.00 & 268.57 & \textbf{272.37} & 0.000 & 16.27 & \textbf{16.98} & 0.000 \\
 & 40000 & 2000 & 10 & 440.00 & 440.00 & 458.53 & \textbf{466.43} & 0.000 & 14.49 & \textbf{19.09} & 0.000 \\
 & 40000 & 2000 & 20 & 440.00 & 440.00 & 458.13 & \textbf{466.30} & 0.000 & 16.04 & \textbf{21.18} & 0.000 \\
 & 40000 & 2000 & 50 & 440.00 & 440.00 & 457.87 & \textbf{466.07} & 0.000 & 16.76 & \textbf{22.72} & 0.000 \\
 & 40000 & 2000 & 100 & 440.00 & 440.00 & 457.60 & \textbf{466.17} & 0.000 & 16.91 & \textbf{23.15} & 0.000 \\
 & 40000 & 4000 & 10 & 440.00 & 440.00 & 455.10 & \textbf{466.33} & 0.000 & 10.31 & \textbf{18.44} & 0.000 \\
 & 40000 & 4000 & 20 & 440.00 & 440.00 & 454.07 & \textbf{466.33} & 0.000 & 11.65 & \textbf{21.13} & 0.000 \\
 & 40000 & 4000 & 50 & 440.00 & 440.00 & 453.93 & \textbf{466.17} & 0.000 & 11.55 & \textbf{22.54} & 0.000 \\
 & 40000 & 4000 & 100 & 440.00 & 440.00 & 453.63 & \textbf{466.27} & 0.000 & 11.43 & \textbf{23.08} & 0.000 \\ \hline
    \end{tabular}
    \end{scriptsize}
\end{table*}

\section{Experimental investigations}
\label{sec5}
We carry out experimental investigations of the coevolutionary Pareto diversity optimization approaches for instances of the maximum coverage problem in graphs which is a well-known NP-hard submodular optimization problem.
Given a weighted graph $G=(V,E)$, with non-negative weights $w(v)$, $v\in V$ on the nodes.
Let $V' \subseteq V$ be a subset of the given nodes and $N(V')$ be the set of nodes that have at least one neighbor in $V'$.
For a given set of nodes $V' \subseteq V$, denote the set of covered nodes by $Cov(V') = V' \cup N(V')$. We identify a given search point $x$ with its set of chosen nodes $V'$ and measure the fitness $f(x) = |Cov(\{v_i \mid x_i=1\})|$ by the number of nodes that are covered by $x$.
Furthermore, we measure the cost $c(x) = \sum_{i=1}^n w(v_i) x_i$ by the sum of the cost of the chosen nodes.

For our experiments, we study costs of nodes that depend on the nodes that they cover. Given a node $v_i$, it covers itself plus all of its neighbours. We set $c(v_i) = (\delta(v_i)+1)^2$ where $\delta(v_i)$ is the degree of node $v_i$ in G. This cost function discourages the choice of nodes of large degrees.

\begin{table*}[]
    \centering
        \caption{Results for improved PDO algorithm variants
 }
    \label{tab:PDO-CH}
      \begin{scriptsize}
    \begin{tabular}{|c|c|c|c||c|c|c|c|c||c|c|c|c|c|c|c|c|}
    \hline
    \multirow{2}{*}{Graph} & \multirow{2}{*}{$B$} & \multirow{2}{*}{$m$} & \multirow{2}{*}{$\mu$} & \multicolumn{5}{|c|}{Optimization} &  \multicolumn{5}{|c|}{Diversity} \\
    
  & &  &  & PDO & PDO-C  & $p_3$ &  PDO-CH  & $p_4$ & PDO & PDO-C   & \normalsize{$p_5$} & PDO-CH   & $p_6$\\ \hline
\multirow{16}{*}{frb30-15-1-mis} & 20000 & 2000 & 10 & 303.60 & \textbf{303.97} & 0.171 & 303.77 & 0.183 & 16.92 & \textbf{18.81} & 0.000 & \textbf{18.81} & 0.853 \\
 & 20000 & 2000 & 20 & 303.70 & 303.83 & 0.882 & \textbf{303.90} & 0.657 & 19.11 & \textbf{21.35} & 0.000 & 21.27 & 0.055 \\
 & 20000 & 2000 & 50 & 303.50 & \textbf{303.87} & 0.274 & \textbf{303.87} & 1.000 & 20.59 & 23.08 & 0.000 & \textbf{23.09} & 0.941 \\
 & 20000 & 2000 & 100 & 303.40 & \textbf{303.87} & 0.169 & 303.77 & 0.506 & 21.10 & 23.70 & 0.000 & \textbf{23.72} & 0.525 \\
 & 20000 & 4000 & 10 & 303.80 & \textbf{303.90} & 0.947 & \textbf{303.90} & 1.000 & 17.97 & 19.44 & 0.000 & \textbf{19.47} & 0.615 \\
 & 20000 & 4000 & 20 & 303.77 & \textbf{303.90} & 0.773 & 303.80 & 0.506 & 20.73 & \textbf{22.25} & 0.000 & \textbf{22.25} & 0.965 \\
 & 20000 & 4000 & 50 & 303.63 & 303.73 & 0.882 & \textbf{303.90}& 0.268 & 23.03 & \textbf{24.32} & 0.000 & \textbf{24.32} & 0.626 \\
 & 20000 & 4000 & 100 & 303.60 & \textbf{303.90} & 0.425 & 303.87 & 0.824 & 23.81 & \textbf{25.17} & 0.000 & \textbf{25.17} & 0.767 \\
 & 40000 & 2000 & 10 & 413.83 & \textbf{415.83} & 0.000 & 415.47 & 0.130 & 24.12 & 29.06 & 0.000 & \textbf{29.09} & 0.767 \\
 & 40000 & 2000 & 20 & 414.10 & 415.70 & 0.000 & \textbf{416.00} & 0.301 & 26.51 & 32.42 & 0.000 & \textbf{32.57} & 0.301 \\
 & 40000 & 2000 & 50 & 413.93 & \textbf{415.83} & 0.000 & 415.70 & 0.762 & 27.86 & \textbf{34.54} & 0.000 & 34.49 & 0.451 \\
 & 40000 & 2000 & 100 & 414.23 & \textbf{416.00} & 0.000 & 415.70 & 0.181 & 28.51 & \textbf{35.36} & 0.000 & 35.19 & 0.084 \\
 & 40000 & 4000 & 10 & 414.17 & \textbf{415.77} & 0.000 & 415.37 & 0.308 & 27.77 & \textbf{31.99} & 0.000 & \textbf{31.99} & 0.684 \\
 & 40000 & 4000 & 20 & 413.90 & \textbf{415.33} & 0.001 & 415.07 & 0.408 & 30.94 & 36.29 & 0.000 & \textbf{36.35} & 0.416 \\
 & 40000 & 4000 & 50 & 413.97 & \textbf{415.30} & 0.000 & 414.97 & 0.359 & 33.63 & 39.39 & 0.000 & \textbf{39.43} & 0.871 \\
 & 40000 & 4000 & 100 & 413.50 & 415.10 & 0.000 & \textbf{415.43} & 0.290 & 34.14 & 40.39 & 0.000 & \textbf{40.41} & 0.668 \\ \hline
 \multirow{16}{*}{frb30-15-2-mis} & 20000 & 2000 & 10 & 289.27 & \textbf{290.00} & 0.000 & \textbf{290.00} & 1.000 & 16.38 & \textbf{17.40} & 0.000 & 17.39 & 0.918 \\
 & 20000 & 2000 & 20 & 289.17 & \textbf{290.00} & 0.000 & \textbf{290.00} & 1.000 & 19.11 & \textbf{20.47} & 0.000 & \textbf{20.47} & 0.790 \\
 & 20000 & 2000 & 50 & 289.07 & \textbf{290.00} & 0.000 & \textbf{290.00} & 1.000 & 21.34 & 22.50 & 0.000 & \textbf{22.51} & 0.112 \\
 & 20000 & 2000 & 100 & 289.23 & \textbf{290.00} & 0.000 & \textbf{290.00} & 1.000 & 22.27 & \textbf{23.29} & 0.000 & \textbf{23.29} & 0.723 \\
 & 20000 & 4000 & 10 & 289.33 & \textbf{290.00} & 0.000 & \textbf{290.00} & 1.000 & 16.47 & \textbf{17.42} & 0.000 & 17.39 & 0.363 \\
 & 20000 & 4000 & 20 & 289.10 & \textbf{290.00} & 0.000 & \textbf{290.00} & 1.000 & 19.29 & 2\textbf{0.50} & 0.000 & 20.49 & 0.530 \\
 & 20000 & 4000 & 50 & 289.17 & \textbf{290.00} & 0.000 & \textbf{290.00} & 1.000 & 21.73 & 22.68 & 0.000 & \textbf{22.71} & 0.329 \\
 & 20000 & 4000 & 100 & 289.07 & \textbf{290.00} & 0.000 & \textbf{290.00} & 1.000 & 22.73 & 23.71 & 0.000 & \textbf{23.73} & 0.433 \\
 & 40000 & 2000 & 10 & 415.10 & \textbf{417.00} & 0.000 & \textbf{417.00} & 1.000 & 25.11 & 29.31 & 0.000 & \textbf{29.43} & 0.174 \\
 & 40000 & 2000 & 20 & 415.47 & \textbf{417.00} & 0.000 & \textbf{417.00} & 1.000 & 27.56 & \textbf{33.11} & 0.000 & 33.00 & 0.478 \\
 & 40000 & 2000 & 50 & 415.57 & \textbf{417.00} & 0.000 & \textbf{417.00} & 1.000 & 29.34 & \textbf{35.55} & 0.000 & 35.52 & 0.584 \\
 & 40000 & 2000 & 100 & 415.63 & \textbf{417.00} & 0.000 & \textbf{417.00} & 1.000 & 30.15 & 36.33 & 0.000 & \textbf{36.36} & 0.647 \\
 & 40000 & 4000 & 10 & 415.37 & \textbf{417.00} & 0.000 & \textbf{417.00} & 1.000 & 26.23 & \textbf{30.18} & 0.000 & 30.05 & 0.065 \\
 & 40000 & 4000 & 20 & 415.03 & \textbf{417.00} & 0.000 & \textbf{417.00} & 1.000 & 29.56 & 34.64 & 0.000 & \textbf{34.73} & 0.294 \\
 & 40000 & 4000 & 50 & 415.67 & \textbf{417.00} & 0.001 & \textbf{417.00} & 1.000 & 31.56 & \textbf{37.44} & 0.000 & 37.41 & 0.535 \\
 & 40000 & 4000 & 100 & 415.17 & 416.90 & 0.000 & \textbf{417.00} & 0.824 & 32.16 & 38.34 & 0.000 & \textbf{38.47} & 0.001 \\ \hline
 \multirow{16}{*}{frb35-17-1-mis}  & 20000 & 2000 & 10 & \textbf{300.00} & \textbf{300.00} & 1.000 & \textbf{300.00} & 1.000 & 14.99 & 16.49 & 0.000 & \textbf{16.55} & 0.105 \\
 & 20000 & 2000 & 20 & \textbf{300.00} & \textbf{300.00} & 1.000 & \textbf{300.00} & 1.000 & 17.15 & \textbf{18.74} & 0.000 & 18.73 & 0.636 \\
 & 20000 & 2000 & 50 & \textbf{300.00} & \textbf{300.00} & 1.000 & \textbf{300.00} & 1.000 & 18.58 & \textbf{20.14} & 0.000 & \textbf{20.14} & 0.337 \\
 & 20000 & 2000 & 100 & \textbf{300.00} & \textbf{300.00} & 1.000 & \textbf{300.00} & 1.000 & 19.05 & \textbf{20.65} & 0.000 & 2\textbf{0.65} & 0.767 \\
 & 20000 & 4000 & 10 & \textbf{300.00} & \textbf{300.00} & 1.000 & \textbf{300.00} & 1.000 & 15.44 & \textbf{16.72} & 0.000 & 16.69 & 0.408 \\
 & 20000 & 4000 & 20 & \textbf{300.00} & \textbf{300.00} & 1.000 & \textbf{300.00 }& 1.000 & 17.93 & 19.04 & 0.000 & \textbf{19.06} & 0.249 \\
 & 20000 & 4000 & 50 & \textbf{300.00} & \textbf{300.00} & 1.000 & \textbf{300.00} & 1.000 & 19.50 & \textbf{20.45} & 0.000 & 20.44 & 0.836 \\
 & 20000 & 4000 & 100 & \textbf{300.00} & \textbf{300.00} & 1.000 & \textbf{300.00} & 1.000 & 19.96 & \textbf{21.00} & 0.000 & 20.96 & 0.231 \\
 & 40000 & 2000 & 10 & 465.20 & \textbf{466.00} & 0.076 & \textbf{466.00} & 1.000 & 25.35 & 28.54 & 0.000 & \textbf{28.58} & 0.615 \\
 & 40000 & 2000 & 20 & 465.40 & \textbf{466.00} & 0.183 & \textbf{466.00} & 1.000 & 28.40 & \textbf{32.86} & 0.000 & 32.81 & 0.379 \\
 & 40000 & 2000 & 50 & 465.57 & \textbf{466.03} & 0.196 & 466.00 & 0.824 & 30.11 & \textbf{35.15} & 0.000 & 35.10 & 0.079 \\
 & 40000 & 2000 & 100 & 464.77 & \textbf{466.03} & 0.005 & 466.00 & 0.824 & 30.72 & \textbf{35.83} & 0.000 & \textbf{35.83} & 0.723 \\
 & 40000 & 4000 & 10 & 465.10 & \textbf{466.03} & 0.018 & \textbf{466.03} & 1.000 & 25.40 & 28.51 & 0.000 & \textbf{28.65} & 0.023 \\
 & 40000 & 4000 & 20 & 465.40 & \textbf{466.07} & 0.058 & \textbf{466.07} & 1.000 & 28.83 & \textbf{33.29} & 0.000 & 33.22 & 0.206 \\
 & 40000 & 4000 & 50 & 465.33 & \textbf{466.03} & 0.132 & \textbf{466.03} & 1.000 & 31.03 & \textbf{35.62} & 0.000 & 35.61 & 0.976 \\
 & 40000 & 4000 & 100 & 465.37 & \textbf{466.00} & 0.121 & \textbf{466.00} & 1.000 & 31.75 & \textbf{36.38} & 0.000 & \textbf{36.38} & 0.965 \\ \hline
 \multirow{16}{*}{frb40-19-1-mis} & 20000 & 2000 & 10 & 272.50 & \textbf{273.00} & 0.001 & \textbf{273.00} & 1.000 & 12.95 & \textbf{13.29} & 0.000 & \textbf{13.29} & 1.000 \\
 & 20000 & 2000 & 20 & 272.50 & \textbf{273.00} & 0.001 & \textbf{273.00} & 1.000 & 15.24 & \textbf{15.87} & 0.000 & \textbf{15.87} & 0.663 \\
 & 20000 & 2000 & 50 & 272.50 & \textbf{273.00} & 0.001 & \textbf{273.00} & 1.000 & 16.55 & 17.34 & 0.000 & \textbf{17.35} & 0.773 \\
 & 20000 & 2000 & 100 & 272.47 & \textbf{273.00} & 0.000 & \textbf{273.00} & 1.000 & 16.97 & 17.78 & 0.000 & \textbf{17.81} & 0.255 \\
 & 20000 & 4000 & 10 & 272.37 & \textbf{273.00} & 0.000 & \textbf{273.00} & 1.000 & 12.96 & \textbf{13.29} & 0.000 & \textbf{13.29} & 0.824 \\
 & 20000 & 4000 & 20 & 272.43 & \textbf{273.00} & 0.000 & \textbf{273.00} & 1.000 & 15.21 & \textbf{15.88} & 0.000 & 15.87 & 0.255 \\
 & 20000 & 4000 & 50 & 272.47 & \textbf{273.00} & 0.000 & \textbf{273.00} & 1.000 & 16.55 & \textbf{17.34} & 0.000 & \textbf{17.34} & 0.813 \\
 & 20000 & 4000 & 100 & 272.37 & \textbf{273.00} & 0.000 & \textbf{273.00} & 1.000 & 16.98 & 17.77 & 0.000 & \textbf{17.80} & 0.344 \\
 & 40000 & 2000 & 10 & 466.43 & \textbf{468.00} & 0.000 & \textbf{468.00} & 1.000 & 19.09 & \textbf{22.98} & 0.000 & 22.97 & 0.647 \\
 & 40000 & 2000 & 20 & 466.30 & \textbf{468.00} & 0.000 & \textbf{468.00} & 1.000 & 21.18 & \textbf{26.46} & 0.000 & 26.37 & 0.169 \\
 & 40000 & 2000 & 50 & 466.07 & \textbf{468.00} & 0.000 & \textbf{468.00} & 1.000 & 22.72 & 28.33 & 0.000 & \textbf{28.34} & 0.487 \\
 & 40000 & 2000 & 100 & 466.17 & \textbf{468.00} & 0.000 & \textbf{468.00} & 1.000 & 23.15 & 28.95 & 0.000 & \textbf{29.00} & 0.128 \\
 & 40000 & 4000 & 10 & 466.33 & \textbf{468.00} & 0.000 & \textbf{468.00} & 1.000 & 18.44 & \textbf{22.98} & 0.000 & 22.92 & 0.442 \\
 & 40000 & 4000 & 20 & 466.33 & \textbf{468.00} & 0.000 & \textbf{468.00} & 1.000 & 21.13 & 26.35 & 0.000 & \textbf{26.40} & 0.387 \\
 & 40000 & 4000 & 50 & 466.17 & \textbf{468.00} & 0.000 & \textbf{468.00} & 1.000 & 22.54 & 28.32 & 0.000 & \textbf{28.35} & 0.359 \\
 & 40000 & 4000 & 100 & 466.27 & \textbf{468.00} & 0.000 & \textbf{468.00} & 1.000 & 23.08 & \textbf{28.98} & 0.000 & \textbf{28.98} & 0.859 \\ \hline
    \end{tabular}
    \end{scriptsize}
\end{table*}

We investigate all algorithms on the graphs frb30-15-1-mis (450 nodes and 17827 edges), frb30-15-2-mis (450 nodes and 17874 edges), frb35-17-1-mis (595 nodes and 27856 edges), frb40-19-1-mis (760 nodes and 41314 edges)  which are benchmarks out of the vertex cover test suite from \cite{datasetsfrb}. The graphs are relatively sparse which makes them well suited for the maximum coverage problem as in dense graphs the choice of a single node can almost cover the whole graph.
We investigate for each graph budgets $B \in \{20000, 40000\}$, margins $m \in \{2000, 4000\}$, and population sizes $\mu \in \{10, 20, 50, 100\}$. Note GDGS is run with a margin $B-m$ and creates $\mu$ individuals. Therefore a larger margin implies a smaller threshold $f_{\min}$ which is determined by the worst function value obtained by GDGS. This allows more room for diversity optimization. 

Each algorithm is run for each of these problem parameter combinations $30$ times where each run is executed for $10$ million fitness evaluations. We intentionally choose such a high budget in terms of fitness evaluation in order to not prematurely stop the optimization process. The tables show values that are the average of the 30 run. For statistical evaluation we use the Mann-Whitney U test and record the $p$-values with $3$ decimal digits. A value of "$0.000$"  in a $p_i$-column indicates a $p$-value less than $0.001$. We call a result significant if the $p$-value is less than $0.05$.

We start by comparing the Pareto Diversity Optimization approach to DIVEA shown in Table~\ref{tab:PDO}. DIVEA obtains its initial population through the generalized diversifying greedy sampling (GDGS) approach as outlined in \cite{DBLP:conf/gecco/NeumannB021}. DIVEA runs GDGS with constraint bound $B-m$ (instead of $B$) and samples for each to be constructed individual additional elements until no further element can be added without violating the constrained bound. 
The fitness of the worst individual obtained by GDGS is used as the quality threshold for the diversity optimization process. GDGS$_w$ and $GDGS_B$ denote for each row the average of the worst and best solution obtained by GDGS within the 30 runs.
DIVEA evolves the population obtained by GDGS and tries to improve the best solution obtained by GDGS and increase the diversity of the solutions in $P_D$.
It has already been shown in \cite{DBLP:conf/gecco/NeumannB021} that DIVEA is able to significantly increase the diversity of the population produced by GDGS. The results in Table~\ref{tab:PDO} also show that the quality of the best solution obtained by DIVEA is substaintially higher than the best solution provided by GDGS. Overall, this shows that the elitist DIVEA approach is able to significantly increase the quality of the best solution and the diversity of the set of solutions of GDGS.

We now consider the optimization result obtained by PDO and compare it to the one obtained by DIVEA. The results show that PDO is able to achieve higher average results for all settings that have been tested. All results are also statistically significant which is evidenced by a $p$-value of less than $0.001$ as shown in the column $p_1$.
Comparing the diversity scores obtained by DIVEA and PDO the same situation can be observed. Again PDO obtains for each setting a higher average diversity score and the results are statistically significant with a $p$-value of less than $0.001$ for almost all instances as shown in column $p_2$.

We now discuss the results for the improved PDO variants PDO-C and PDO-CH shown in Table~\ref{tab:PDO-CH}. The table also lists the PDO results again to make the comparison easier.
PDO-C use the inter-population crossover and mutation approach outlined in Algorithm~\ref{alg:crossmut} but still uses standard bit mutations as PDO for mutation. PDO-CH differs from PDO-C by using heavy tail mutations instead of standard bit mutations.
We use $p_c=0.2$ for both approaches as the Pareto optimization part based on mutations is crucial for the optimization and we expect that crossover will provide additional benefit through mixing individuals. When we apply heavy tail mutation, we use $\beta=1.5$ which is a standard recommended setting.

It can be observed that the improved variants can slightly improve on the optimization result for PDO for some of the investigated settings while showing no significant difference for the graphs frb30-15-1 and frb35-17-1 when using cost constraint $B=20000$. Statistical test results for PDO-C and PDO are shown in column $p_3$.
Furthermore PDO-C and PDO-CH show very similar average scores for optimization. Statistical tests results comparing the optimization results of PDO-C and PDO-CH displayed in column $p_4$ do not show a significant difference in terms of all considered settings. This leads us to the conclusion that there is no benefit of the used heavy tail mutation operator. We think that this is not too surprising as the standard Pareto optimization approach has already shown to be very successful when using standard bit mutations and the use of crossover may also limit the additional benefit of heavy tail mutations.

Examining the diversity results of PDO-C and PDO-CH it can be observed that both algorithms achieve a higher average diversity score for all settings than PDO. All statistical tests carried out for PDO-C and PDO shown under column $p_5$ show statistical significance with a $p$-value less than $0.001$. Comparing PDO-C and PDO-CH, there is no clear superior approach. Average diversity scores are very similar and statistical results shown in column $p_6$ comparing the diversity results of PDO-C and PDO-CH show no significant difference.
\section{Conclusions}
Computing high quality diverse set of solutions provides decision makers with different options of implementing a high quality solution. The area has obtained increasing interest in the evolutionary computation literature in recent years. We introduced Pareto Diversity Optimization which is a coevolutionary approach optimizing the quality of the best possible solution as well as computing a diverse set of $\mu$ solutions meeting a given threshold value. Our experimental results show that PDO is outperforming the previously introduced DIVEA algorithm both in terms of quality and the diversity of the obtained solutions for the submodular maximum coverage problem. Furthermore, we have shown that the use of inter-population based crossover further increase the diversity of the set of solutions.

\section{Acknowledgements}
This work has been supported by the Australian Research Council (ARC) through grants DP190103894, FT200100536, and by the South Australian Government through the Research Consortium “Unlocking Complex Resources through Lean Processing”.

\bibliographystyle{unsrt}
\bibliography{references}

\end{document}